\documentclass[letterpaper, 10 pt, conference]{ieeeconf}  

\IEEEoverridecommandlockouts                              

\usepackage{cite}
\usepackage{amsmath,amssymb,amsfonts}
\usepackage{algorithmic}
\usepackage{graphicx}
\usepackage{textcomp}
\usepackage{xcolor}
\usepackage{booktabs}
\usepackage{tabularx}
\usepackage{hyperref} 

\title{\LARGE \bf
An Empirical Evaluation of Neural and Neuro-symbolic Approaches to Real-time Multimodal Complex Event Detection
}

\author{Liying Han and Mani Srivastava$^{1}$
\thanks{$^{1}$Liying Han and Mani Srivastava are at the Electrical and Computer Engineering Department, the University of California, Los Angeles, US, {\tt\small liying98@ucla.edu} and {\tt\small mbs@ucla.edu}}%
\thanks{*Codes are available at \url{https://github.com/nesl/CED_Methods_Eval}.}%
}

\begin{document}

\maketitle
\thispagestyle{empty}
\pagestyle{empty}

\begin{abstract}
For robots and other autonomous cyber-physical systems to have a situational understanding of their environments and to interact with humans, they need to understand \textit{complex events (CEs)} using observations from their sensors. While end-to-end neural architectures can effectively process noisy and uncertain sensor data, their limited context sizes and reasoning abilities restrict them from identifying complex events that unfold over long intervals of time and space with complicated relationships among multiple instantaneous and local events. Recently, \textit{Complex Event Processing} systems based on \textit{neuro-symbolic} methods combine neural models with symbolic models based on human knowledge. By bootstrapping on human knowledge, they seek high performance with limited training data. However, a full understanding of the relative capabilities of these two approaches to complex event detection (CED) in performing temporal reasoning is missing. We explore this gap by investigating in depth the performance of various neural and neurosymbolic architectures. Specifically, we formulate a multimodal CED task to recognize \textit{CE} patterns in real-time from IMU and acoustic streams. We evaluate (i) \textit{end-to-end neural} architectures that detect \textit{CE} directly from the sequence of sensor embedding vectors using various neural backbone models, (ii) two-stage \textit{concept-based neural} architectures that first map the sequence of sensor embedding vectors to a sequence of \textit{atomic events (AEs)} with a fully-connected neural model and detect \textit{CE} using various neural backbone models, and (iii) a \textit{neuro-symbolic} architecture that uses a symbolic finite-state machine to detect a \textit{CE} from a sequence of \textit{AE}s. Our empirical analysis indicates the neuro-symbolic approach consistently outperforms purely neural architectures by a significant margin, even in settings where the neural models were trained using a large training set of \textit{CE} examples and had access to sufficient input temporal context to detect the entire \textit{CE}. 
\end{abstract}

\section{Introduction}
In scenarios where real-time human-robot interaction is necessary, such as social robots and autonomous driving, it is crucial for robots to be capable of detecting and responding to \textbf{complex events (\textit{CE}s)}. Complex events involve intricate timelines, unpredictable durations, and various subevent combinations. They offer high-level contextual information about the environment, enabling robots to provide better feedback or assistance to humans. For in-home robots, such as the SONY AIBO robotic dog \cite{AIBO}, which provides personalized companionship, and the Vayyar Care \cite{Vayyar}, which monitors bathroom usage and detects falls for the elderly, complex events centered around humans are even more important.

However, identifying complex events can be challenging. For instance, ``\textit{irregular bathroom habits}" referring to excessive bathroom use may indicate health issues like frequent urination or diarrhea. To detect such events, a system must match real-time event patterns from noisy sensor streams collected by local or public sensors or sensors placed on the human body. It must recognize the behavior of ``using the bathroom" and ignore irrelevant events while keeping track of the frequency and timing of human behavior that reflects ``irregularity". Systems need temporal reasoning ability to analyze complex events over time, methods only identifying instantaneous events are insufficient to achieve this task. 

\begin{figure}[tbp]
\centerline{\includegraphics[width=1\columnwidth]{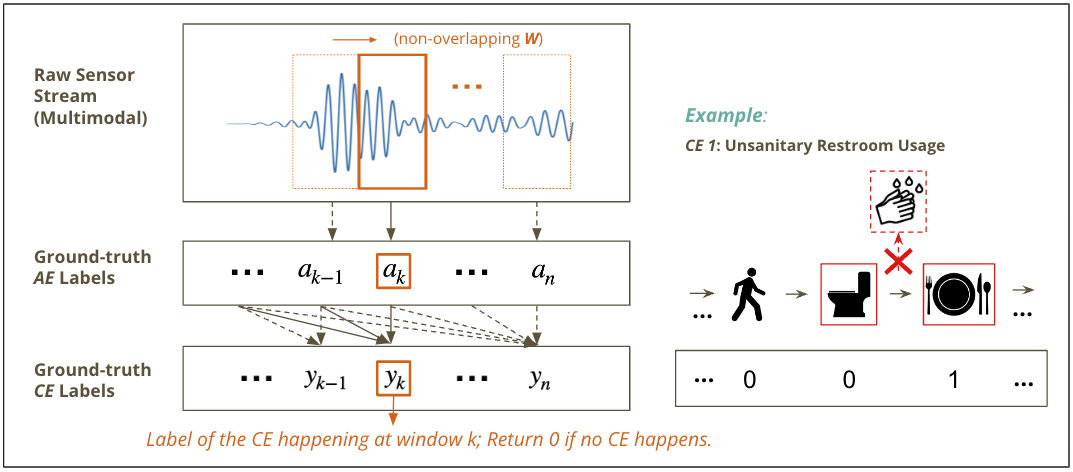}}
\caption{An illustration of the real-time complex events detection task. The example on the right shows that ``\textit{Using Restroom}" and ``\textit{Eating}" without ``\textit{Washing hands}" triggers the complex event detection, but only at the last action ``\textit{Washing hands}" we attach the \textit{CE} label ``1" of this complex event.}
\label{fig:ced-illustration}
\end{figure}

While end-to-end neural architectures can effectively process noisy and uncertain sensor data, their limited context sizes and reasoning abilities restrict them from identifying complex events. Recent \textbf{Complex Event Processing (CEP)} systems \cite{ROIGVILAMALA2023119376} \cite{xing2020neuroplex} \cite{mao2021TOQ} based on neuro-symbolic methods combine data-driven neural models with symbolic models based on human knowledge. However, there is a lack of thorough comparison of these two approaches to complex event detection in performing temporal reasoning over noisy sensor data.

To address this question, we explore in depth the performance of various neural and neurosymbolic architectures. Specifically, we formulate a multimodal CED task to recognize \textit{CE} patterns in real-time using IMU and acoustic streams. We evaluate (i) \textit{end-to-end neural} architectures, (ii) two-stage \textit{concept-based neural} architectures, and (iii) a \textit{neuro-symbolic} architecture on our synthesized 5-min \textit{CE} dataset. Our empirical analysis indicates the neuro-symbolic approach consistently outperforms purely neural architectures by $41\%$ on the average $F1$ score, even when the neural models were trained using a large training set of \textit{CE} examples, and had context sizes larger than the length of the longest temporal \textit{CE} patterns to detect the entire \textit{CE}.

\section{Related Work}
\textbf{Complex event processing.} 
CEP was initially widely studied in traditional stream processing for databases, where incoming data are structured with user-defined features. Early works including \cite{CEPoverview_2012}, \cite{Schultz_2009}, and \cite{Debar_2001} design systems to process complex events via symbolic algorithms such as SQL operations. Those systems analyze structured data with an event engine to identify complex events with specific patterns. For example, fraud detection uses continuous streams of credit card transactions, and intrusion detection systems analyze network traffic in real time to generate alerts and prevent something unexpected. However, as systems nowadays process raw sensory data that are noisy, high-dimensional, and unstructured, CEP with purely symbolic architectures becomes inefficient and fragile to errors and missing data.

\textbf{CEP with neurosymbolic methods.} 
Neurosymbolic CEP systems combine the analysis ability of NNs with the reasoning ability of symbolic components. Neuroplex \cite{xing2020neuroplex} trains an NN module using a CEP engine for knowledge distillation, and DeepProbCEP \cite{ROIGVILAMALA2023119376} uses DeepProbLog \cite{NEURIPS2018_dc5d637e} \cite{DBLP:journals/corr/abs-1907-08194}, a probabilistic logic programming framework, to make CEP rules differentiable. Both systems are end-to-end trainable, but only consider complex events with simple timelines and patterns. A recent study \cite{li-2021-future} has induced complex event schemas from natural language texts by embedding spatial and temporal event structures into Graph Neural Network models \cite{4700287}. However, their complex events are limited to a conceptual level and do not have real-time features. Our work focuses on the real-time detection of complex events with longer and more intricate temporal relations between events.

\textbf{Multimodal Compex event datasets.}
Complex event datasets for evaluating long-term memory and reasoning ability are limited. While there are benchmarks \cite{Tay2020LRA} \cite{bai2023longbench} for long context reasoning, they mainly focus on long text reasoning in NLP tasks. Therefore, we have created a multimodal complex event dataset with IMU and audio sensor data. To generate this dataset, we designed a stochastic \textit{CE} simulator that combines multimodal sensory data from short activities in different ways. It is worth noting that there are several multimodal human activity datasets \cite{Frade-2008-9926} \cite{martínezzarzuela2023vidimu}, but we still synthesized our own data for short activities since their activity classes are not suitable for building our complex events.

\section{Complex Event Detection Task Formulation}

\subsection{Definitions}
\textit{\textbf{Atomic Events (AEs)} are low-level, short-time, and the smallest building blocks of complex events.} Usually, they are instantaneous events that most popular deep learning models can easily recognize. For example, an image classification, a 2-second audio clip classification, or human activity visual recognition.

\textit{\textbf{Complex Events (CEs)} are higher-level events defined by atomic events following some temporal patterns.} Suppose we have a set of atomic events $A$ with $n$ elements $A = \{a_1, a_2, \ldots, a_n\}$. Additionally, we have a set of complex events of interest with $k$ elements $E = \{e_1, e_2, \ldots, e_k\}$. Each complex event $e_i$ is defined by a subset of atomic events $A_i$, following some specific rules:
\begin{align}
    &e_i = R_i(A_i) = R_i(a_i^1, a_i^2, \ldots, a_i^{n_i}),\nonumber\\ &\textrm{where }a_i^j \in A_i, \quad 1 \leq j \leq n_i, \quad 1 \leq i \leq k.
\end{align}
Here, $R_i$ represents the temporal pattern function that maps the relevant atomic events into a complex event $e_i$, $A_i \subseteq A$, and $n_i$ is the number of atomic events in the subset $A_i$.

\subsection{Complex Event Detection Task}
Without loss of generality, let's assume a system receives raw data streams $\mathbf{X_1}$ and $\mathbf{X_2}$ from two sensors with different modalities $M_1$ and $M_2$, respectively. These sensors have distinct sampling rates - the sampling rate is $r_1$ for $M_1$ sensor and $r_2$ for $M_2$ sensor. The system processes the data streams using a non-overlapping sliding window with a fixed length $\Delta t$. At the $t$th sliding window, it gets a multimodal sensor data pair:
\begin{equation}
    \mathbf{D}_t=\left(\mathbf{X_1}(t), \mathbf{X_2}(t)\right),
\end{equation}
where $\mathbf{X_1}(t) \in \mathbb{R}^{(r_1 \times \Delta t) \times m_1}, \mathbf{X_2}(t) \in \mathbb{R}^{(r_2 \times \Delta t) \times m_2}$, with $m_1$ and $m_2$ being the feature dimensions of sensors data from $M_1$ and $M_2$.

Also, at each sliding window $t$, there exists a corresponding ground-truth \textit{CE} label $y_t$, representing the complex event that happens now. As shown in Fig.~\ref{fig:ced-illustration}, $y_t$ depends on all the \textit{AE}s that occurred in previous $t-1$ windows and the current window $t$, and the example on the right shows the real-time \textit{CE} labeling manner we adopt. Suppose the full pattern of a complex event spans from $t_1$ to $t_2$, and then only $y_{t_1}$ is the corresponding \textit{CE} label, all \textit{CE} labels from $y_{t_1}$ to $y_{t_2 - 1}$ are ``0"s, indicating that no complex event is detected before $t_1$. Then for data streams with up to $T$ sliding window clips, the objective of the robot with a real-time CED system $f$ is to accurately predict the complex event label at each sliding window $t$, i.e., to minimize the difference between the predicted $CE$ label $\hat{y_t}$ and the ground-truth $CE$ label ${y_t}$:
\begin{equation}\label{eq:1}
    \min |\hat{y_t} - y_t|, \quad \textrm{ where }\hat{y_t} = f\left(\mathbf{D}_t\right), \quad 1 \leq t \leq T,
\end{equation}
which is a \textit{\textbf{multi-label multi-class classification}} problem. Let $\mathbf{y} = {y_1, y_2, \ldots, y_T}$ be the ground-truth complex event label sequence. Equation~\ref{eq:1} can be expressed in vector form as
\begin{equation}
    \min ||f(\mathbf{D}) - \mathbf{y}||, \quad \textrm{ where } \mathbf{D} = \left(\mathbf{X_1}, \mathbf{X_2}\right).
\end{equation}

\section{Multimodal Complex Event dataset}\label{sec:dataset}

\subsection{Designing Complex Events for In-home Robots}
We design a complex event dataset for in-home healthcare robots. Similar to the Vayyar Care device that uses mmWave sensing due to privacy concerns \cite{Vayyar}, our dataset adopts the optics-free idea and considers audio and IMU data collected on humans shared with robots to detect complex events that indicate bad hygiene habits in real time.

We propose three complex events, ``\textit{Unsanitary Restroom Usage}" ($e_1$), ``\textit{Unsanitary Diet Habit}" ($e_2$), and ``\textit{Bad Brushing Habit}" ($e_3$), with user-defined patterns, and one default class $e_0$ if none of the above happens, as shown in Table~\ref{tab:complex-event}. Note that if an event does not belong to classes $e_1, e_2$, or $e_3$, it is categorized as the default event $e_0$.
\begin{table*}[ht]
\caption{Definition of complex event classes
    }
    \centering
    \begin{tabularx}{\textwidth}{l X}\toprule
    Complex Events & Definition\\\midrule
    No Complex Events ($e_0$) &  No interesting complex events are detected. The default class if none of the following complex events happen.\\
    Unsanitary Restroom Usage ($e_1$)  & After using the restroom, no washing hands followed by other activities, or walking away for more than 1 minute.\\
    Unsanitary Diet Habit ($e_2$)  & No hand washing before meals. The maximum time gap between washing hands and eating is 2 minutes.\\
    Bad Brushing Habit ($e_3$) & Brushing teeth for less than 2 minutes. If no teeth brushing occurs within 10 seconds, it is considered stopped.\\
    \bottomrule
    \end{tabularx}
    
    \label{tab:complex-event}
\end{table*}

\subsection{Complex Event Simulator}
Due to the complexity of complex event patterns, each \textit{CE} has infinitely many combinations of \textit{AE}s over time. To generate a general distribution for complex events, we developed a stochastic \textit{CE} human activity simulator to synthesize multimodal time-series data for each \textit{CE} pattern. 

\begin{figure}[tbp]
\centerline{\includegraphics[width=1\columnwidth]{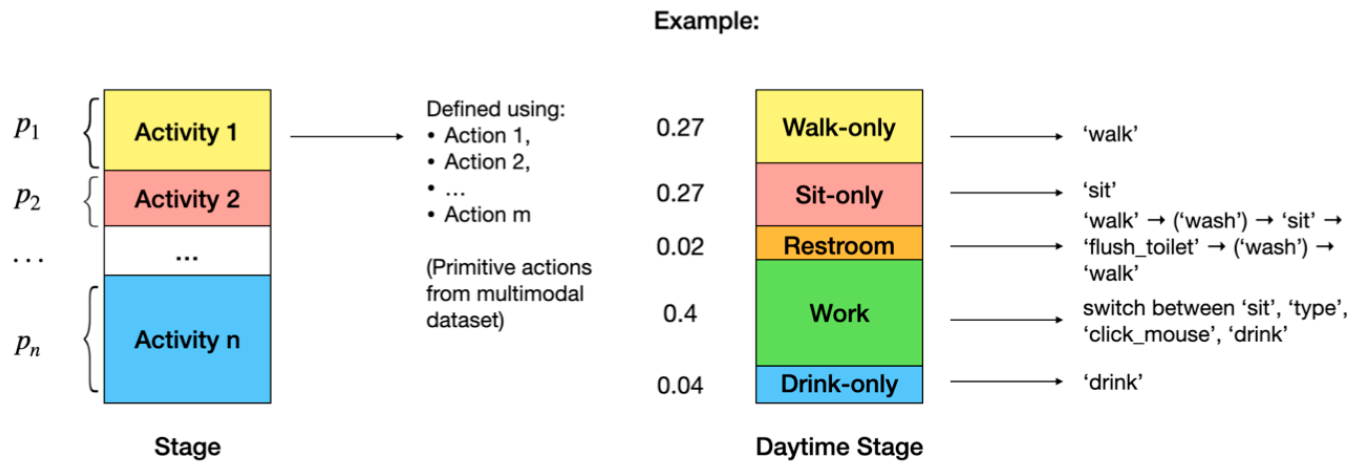}}
\caption{\textbf{Daily activity simulator.} Each \textit{Stage} has a set of $n$ \textit{Activities} that may happen according to a predefined distribution, where \textit{Activity} $i$ has a probability $p_i$ of taking place in that \textit{Stage}. Each \textit{Activity} is defined by a temporal combination of relevant \textit{AE}s. For example, in \textit{Daytime Stage} \textit{Activities} ``\textit{Walk-only}", ``\textit{Sit-only}", ``\textit{Restroom}," ``\textit{Work}", and ``\textit{Drink-only}" happen with probabilities $[0.27, 0.27, 0.02, 0.4, 0.04]$ respectively. Each \textit{Activity} is defined by the pattern displayed on the right side.}
\label{fig:ce-simulator}
\end{figure}

Fig.~\ref{fig:ce-simulator} shows the simulator we implemented to generate stochastic \textit{CE} sequences. It has multiple \textit{Stages}, and each \textit{Stage} has a set of \textit{Activities} that happen with different probabilities. An \textit{Activity} is defined by a temporal pattern of \textit{Actions}. For example, the \textit{Activity} ``\textit{Use Restroom}" is defined as the sequential combination of \textit{Actions}: \textit{`walk' $\rightarrow$ (`wash') $\rightarrow$ `sit' $\rightarrow$ `flush-toilet' $\rightarrow$ (`wash') $\rightarrow$ `walk'}, where the parentheses indicate that the \textit{Action} inside only happens with certain probability. The duration of each \textit{Action} is also random within some user-defined threshold. Hence, we can get sequences for

We synthesize the multimodal sensor data of 9 \textit{Actions}, which also serve as the \textit{AE}s to build \textit{CE}s. Each \textit{Action} class corresponds to a pair of audio and IMU classes, as shown in the first column of Fig.~\ref{tab:multimodal-action}. The 9 audio classes and 9 IMU classes are selected from the following two datasets:

\subsubsection{Audio dataset}
We utilize the ESC-70 dataset, a combination of the ESC-50 dataset \cite{piczak2015dataset} and the Kitchen20 dataset \cite{8877429}. The ESC-50 dataset consists of 2000 5-second labeled environmental audio recordings, with 50 different classes of natural, human, and domestic sounds. The Kitchen20 dataset collects 20 labeled kitchen-related environmental sound clips. We also self-collected 40 additional 5-second silent sound clips for \textit{Actions} that do not have sound. We downsampled the original sampling rate of those recordings is 44.1 kHz to 16 kHz.

\subsubsection{IMU dataset} 
We use the WISDM dataset \cite{8835065}, containing raw accelerometer and gyroscope sensor data collected from smartphones and smartwatches, at a sampling rate of 20 Hz. It was collected from 51 test subjects as they performed 18 activities for 3 minutes each. Based on the findings in a survey paper \cite{oluwalade2021human}, We only use smartwatch data for better accuracy. We segment the original data samples into non-overlapping 5-second clips.

Using those two datasets, we generate 500 multimodal data samples of 5 seconds per \textit{Action} class. To synthesize multimodal \textit{CE} sensor data, the 5-second sensor data clips for every \textit{Action} are concatenated according to the \textit{AE} patterns of the generated stochastic \textit{CE} sequences.

\subsection{Generating Ground-truth CE Labels}
To create the multimodal \textit{CE} dataset, we still need the ground-truth \textit{CE} labels. We implemented a finite state machine (FSM) for every complex event class, following the \textit{CE} rules defined in Table~\ref{tab:complex-event}. For each \textit{CE} sequence generated stochastically by the \textit{CE} simulator, we pass the corresponding sequence of ground-truth \textit{AE} labels to the FSMs and let them do the real-time \textit{CE} class labeling automatically.

\begin{table}[htbp]
\caption{Definition of multimodal action classes}
\setlength\tabcolsep{1.5pt}%
    \centering
    \begin{tabular}{ l  r  r }\toprule
    \textbf{Multimodal \textit{Action}} & \textbf{Audio class} & \textbf{IMU class} \\\midrule
    \textbf{walk} & footsteps & walking \\
    \textbf{sit} & no sound & sitting \\
    \textbf{brush teeth} & brushing teeth & teeth \\
    \textbf{click mouse} & mouse click & sitting \\
    \textbf{drink} & drinking sipping & drinking \\
    \textbf{eat} & eating & eating pasta \\
    \textbf{type} & keyboard typing & typing \\
    \textbf{flush toilet} & toilet flush & standing \\
    \textbf{wash} & water-flowing & standing \\
    \bottomrule
    \end{tabular}
    \label{tab:multimodal-action}
\end{table}

\section{Real-time Complex Event Detection System}
\subsection{Overview}
We propose a two-module real-time processing system to facilitate a comprehensive comparison between neural-only and neuro-symbolic approaches on the CED task. As is shown in the left of Fig.~\ref{fig:system-overview}, the system takes both audio and IMU streams as inputs and applies a non-overlapping sliding window of size $W$, which is 5 seconds, to process data streams into segments. Those segments are forwarded to the following two modules:

\begin{figure*}[tb]
    \centering
    \includegraphics[width=1\textwidth]{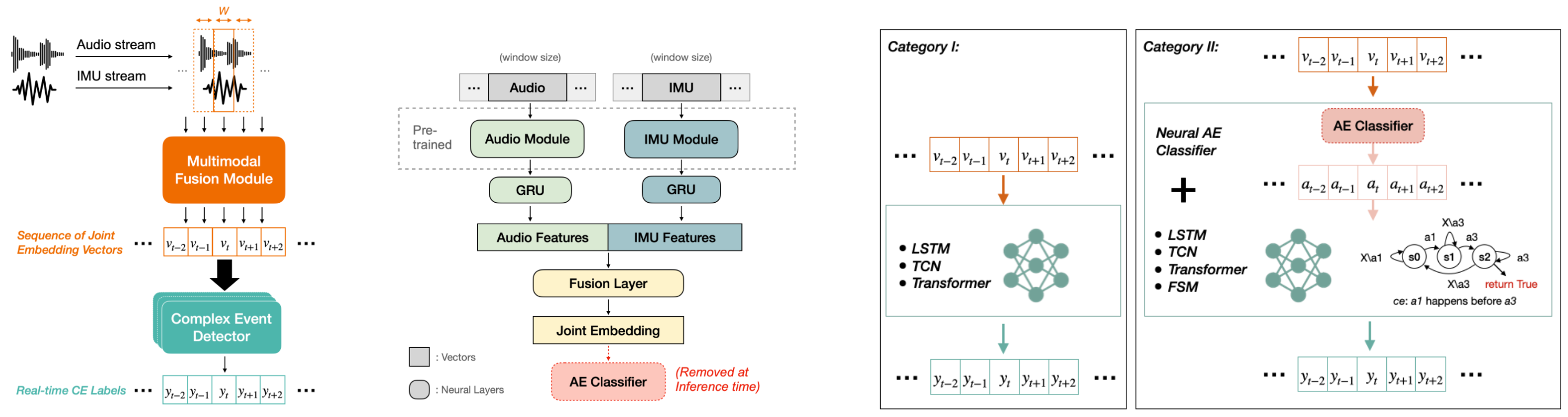}
    \caption{Overview of (\textbf{Left}) the entire real-time CED system, (\textbf{Middle}) the multimodal fusion module, and (\textbf{Right}) the complex event detector module.}
    \label{fig:system-overview}
\end{figure*}

\subsubsection{Multimodal Fusion Module}
This component encodes each window segment $s_i$ into a joint embedding vector $v_i$, concatenating them into a time sequence $\{v_i\}_{i=1}^T$ for downstream tasks. It is used to efficiently utilize information from multiple sources with varying sampling rates of sensor data. It minimizes the impact of different sensor data feature extraction capabilities on the comparison results by ensuring that all neural-only and neural-symbolic models receive the same data embedding processed by this module.

\subsubsection{Complex Event Detector}
It takes the embedding vectors $\{v_i\}_{i=1}^T$ to detect \textit{CE} patterns. Its output is a real-time \textit{CE} label sequence $\{y_i\}_{i=1}^T$, where $y_i$ is \textit{CE} label at window $i$. This detector has neural-only and neuro-symbolic alternatives.

Denote the multimodal fusion module as $m$, and the complex event detector as $g$, the CED task objective defined in Eq.~\ref{eq:1} becomes:
\begin{equation}
    \min |\hat{y_t} - y_t|, \quad \textrm{ where } \hat{y_t} = g\left(m\left(\mathbf{D}_t\right)\right), \quad 1 \leq t \leq T,
\end{equation}
The multimodal fusion module $m$ remains consistent, while the \textit{CE} detector $g$ varies for comparison among models.
 
\subsection{Multimodal Fusion Module}
\subsubsection{Early Fusion Model}
 Various approaches can be considered for multimodal, such as early fusion, late fusion, and hybrid fusion \cite{DBLP:journals/corr/BaltrusaitisAM17}. In our system, we employ early fusion, which involves integrating the features from both modalities immediately after extraction. The middle of Fig.~\ref{fig:system-overview} provides an overview of the multimodal fusion module's structure. First, the model uses pre-trained audio and IMU modules, BEATs and LIMU-Bert, respectively, to extract features from every 5-second audio and IMU clip. Then the audio and IMU embeddings undergo individual Gated Recurrent Unit (GRU) layers to obtain audio and IMU embeddings of the same dimension 128. Next, we concatenate them into an embedding of 256 and pass it through a Fusion Layer to create a joint representation of 128, which is trained alongside a downstream fully-connected NN for \textit{AE} classification. Notably, during the inference phase, the classifier is omitted, and only the output from the last hidden layer of the fusion layer is employed as the fusion embedding. 
\subsubsection{Training}
The BEATs model used for the audio module is already pre-trained, while the LIMU-Bert model for IMU module is pre-trained on large datasets. we freeze the parameters of both models and focus on training the other components of the multimodal fusion module. For training and testing the fusion module, we utilize the multimodal atomic action dataset introduced in Section~\ref{sec:dataset}. The training process minimize the cross-entropy loss:
\begin{equation}
    \min_\theta L_{m}= - \min_\theta \frac{1}{N}\left(\sum_{i=1}^N c_i \cdot \log \left(\hat{c_i}\right)\right), \textrm{where } \hat{c_i} = m_\theta(\mathbf{d_i})
\end{equation}
where $\mathbf{d}$ is the multimodal sensor data of 5-second window size, $\hat{c_i}$ is the predicted label of the \textit{AE} in that window, $c_i$ is the corresponding ground truth label, and $N$ represents the size of the multimodal atomic action dataset.

\subsection{CE Detector}
\subsubsection{Requeirements}
We design various NN and NS architectures to detect complex events in real time. They satisfy the following components:

\textbf{Causal structure.}
Real-time CED requires the system to predict a complex event that occurs at every time step $t$, so only the knowledge of events from previous timestamps (0 to $t-1$) should be used to identify the complex event occurring at the current time. The models should have causal structures that restrict them from accessing future information.

\textbf{Minimum receptive field.} 
The receptive field/context size of each model should be larger than the longest temporal relations of complex events, to ensure that the model can at least ``see" the full pattern of complex events.

\subsubsection{Alternatives}
We design three architectures that fall into the two categories in the right part of Fig.\ref{fig:system-overview}.

\textbf{End-to-end Neural Architecture.} This architecture belongs to \textit{Category I}, where neural models directly take the 128-dimensional sensor embedding vectors of the multimodal information for each window to detect complex events. We choose (1) a \textbf{\textit{Unidirectional LSTM}} \cite{hochreiter1997long} of 3 LSTM layers with a hidden dimension of 128; (2) a \textbf{\textit{Causal TCN}}\cite{bai2018tcn} that masks information from future timestamps in convolutional operations. It has one stack of residual blocks with a last dilation rate of 16 and a kernel size of 2. The number of filters for each level is 256. The receptive field is calculated as 
$\textrm{\# stacks of blocks} \times \textrm{kernel size} \times \textrm{last dilation rate}= 1 \times 2 \times 16 = 32$, which is greater than the longest 2-min temporal pattern of our \textit{CE} dataset, corresponding to a receptive field greater than $ 2 \times 60 \div 5 = 24$ (\# seconds divided by the window size); and (3) A \textbf{\textit{Causal Transformer Encoder}} with a triangular attention mask to restrict the model's self-attention to previous timestamps only, excluding information from future timestamps. The causal transformer encoder uses 6 encoder layers with a hidden dimension 128, multi-head attention with 8 heads, and positional encoding \cite{DBLP:journals/corr/VaswaniSPUJGKP17}.

\textbf{Two-stage Concept-based Neural Architecture.} This architecture falls into \textit{Category II}. It first maps the sequence of sensor embedding vectors to a sequence of \textit{AE}s with a fully-connected neural model and detects \textit{CE} using various neural backbone models, denoted as \textbf{\textit{Neural AE + X}}. Here the \textit\textbf{{X}} stands for the backbone models, which are \textbf{\textit{LSTM}}, \textbf{\textit{TCN}} and \textbf{\textit{Transformer}} with almost the same hyperparameters of previous models in {End-to-end Neural Architecture}, except that they take the 9-dimensional one-hot vectors as \textit{AE} concepts.

\textbf{Neuro-symbolic Architecture.} This architecture also belongs to \textit{Category II}. The neuro-symbolic model we choose uses a user-defined symbolic FSM to detect a \textit{CE} from a sequence of \textit{AE}s.

\subsubsection{Training NN Baselines}
We encounter the class imbalance issue when training models with an end-to-end neural architecture and a two-stage concept-based neural architecture. Due to the temporal sparsity of \textit{CE} labeling, the ground-truth label sequence is almost all ``0"s, which is very similar to the issue in object detection when the number of negative classes (background objects) overwhelms the number of positive classes (foreground objects). Hence, we adopt Focal Loss (FL) \cite{DBLP:journals/corr/abs-1708-02002}, a variant of the cross-entropy loss that penalizes more heavily on mistakes made on less frequent class examples. Here we only consider a binary classification problem for notation simplicity. 
Given a complex event dataset with $N$ training examples, and each example sequence $y_i$ has length $T$, Let $p_y$ be the estimated probability of \textit{CE} class to be $y$, the training objective is to optimize the total FL loss:
\begin{equation}
\min_\theta L_{CE}\left(\theta\right)=-\sum_{i=1}^{N}\sum_{t=1}^{T}\alpha_{y_i(t)}\left(1-p_{y_i(t)}\right)^\gamma \log \left(p_{y_i(t)}\right).
\end{equation}
where $\gamma$ is the focusing parameter that smoothly adjusts the rate at which frequent classes are down-weighted, and $\alpha_{y}$ is the class weight coefficient. We choose $\gamma = 2$ as recommended, $\alpha_0 = 0.005 $ for the most frequent \textit{CE} class ``0", and the class weights for other classes are set to $\alpha_y = 0.45$.

\section{Experiments}
\subsection{Experimental Setup}
We set the window size to be 5 seconds for the CED system. For training, we utilize two Nvidia GPUs (1080-Ti). The multimodal fusion module is trained for 100 epochs, while all the NN models for the \textit{CE} Detector are trained for 2000 epochs. We use the SGD optimizer with a learning rate of $1e-3$ for all training processes. We randomly chose 10 random seeds for each training episode.
\subsection{Evaluation}
\subsubsection{Dataset}
We use the synthesized datasets described in Sec.~\ref{sec:dataset}. We use the multimodal \textit{Action} dataset in Tab.~\ref{tab:multimodal-action} to train the multimodal fusion module. We generated 500 training samples and 100 test samples. To train the neural \textit{CE} detectors, we synthesized 5-min complex events using the \textit{CE} dataset in Tab.~\ref{tab:complex-event} with our \textit{CE} Simulator, so each complex event example has 60 windows ($5 ~min \times 60~sec/min \div 5 ~sec= 60$). We generate 10,000 training and 1,000 test examples for the multimodal \textit{CE} dataset.
\subsubsection{Metrics}
We use \textit{precision}, \textit{recall}, and \textit{F1} score as the performance metrics. As the \textit{CE} label sequences contain many too ``0"s (\textit{CE} class $e_0$) that we do not care about, $e_0$ is considered as a negative class, and $e_1$, $e_2$ and $e3$ are positive classes. 
Given the number of true positives ($tp$), false positives ($fp$), and false negatives ($fn$), we have
\begin{equation}
 { precision } =\frac{t p}{t p+f p},
\quad 
 { recall } =\frac{t p}{t p+f n},
\end{equation}
and $F1$ score is the harmonic mean of \textit{precision} and \textit{recall}.
A higher \textit{precision} score reflects the model's ability to correctly classify positive samples, while a higher \textit{recall} score reflects the model's ability to identify more positive samples.

\subsection{Results \& Analysis}
\subsubsection{Evaluation of the \textit{AE} classifier} We test the classifier using the multimodal \textit{AE} dataset, which is used in Neural \textit{AE} + X models. The classifier achieves 91\% accuracy on test set.

\subsubsection{Evaluation of CE detector alternatives}
The evaluation results of NNs and NS models on the real-time complex event detection task are shown in Tab.~\ref{tab:results}. We calculate metric scores for each \textit{CE} class and their averages, and also the average $F1$ scores for all \textit{CE} classes and average $F1$ scores for only positive \textit{CE} classes. 

\begin{table*}[tb]
\caption{\textit{precision}, \textit{recall} and $F1$ scores with 95\% CI (with NNs trained on 10000 examples)}
    \begin{center}
    \setlength\tabcolsep{1.5pt}%
    \begin{tabular}{@{}lccccccccccccccc@{}}\toprule 
    & \multicolumn{5}{c}{$precision$} & & \multicolumn{5}{c}{$recall$}  && \multicolumn{2}{c}{$F1$}\\
    \cmidrule{2-6} \cmidrule{8-12} \cmidrule{14-15}
    & $e_0$ & $e_1$ & $e_2$ & $e_3$ & Avg. && $e_0$ & $e_1$ & $e_2$ & $e_3$ & Avg.&& All & Pos.\\ \midrule
    LSTM & .98 $\pm$ 0& 0 $\pm$ 0& 0 $\pm$ 0& 0 $\pm$ 0& .25 $\pm$ 0&& \textbf{1.0} $\pm$ 0& 0 $\pm$ 0& 0 $\pm$ 0 & 0 $\pm$ 0& .25 $\pm$ 0 && .25 $\pm$ 0 & 0 $\pm$ 0\\
    TCN & .997 $\pm$ .003 & .19 $\pm$ .08 & .25 $\pm$ .05 & .07 $\pm$ .003 & .38 $\pm$ .02 && .90 $\pm$ .03 & .54 $\pm$ .58 & \textbf{1.0} $\pm$ 0 & .94 $\pm$ .04 & .84 $\pm$ .13 && .43 $\pm$ .04 & .26 $\pm$ .06\\
    Transformer & .99 $\pm$ .002 & .02 $\pm$ .01 & .51 $\pm$ .12 & .10 $\pm$ .02 & .41 $\pm$ .03 && .92 $\pm$ .08 & .24 $\pm$ 63 & \textbf{1.0} $\pm$ 0 & .94 $\pm$ .02 & .77 $\pm$ .15 && .46 $\pm$ .02 & .30 $\pm$ .04 \\\midrule
    Neural AE \\
    \phantom{A} + FSM & .996 $\pm$ 0 & \textbf{.79} $\pm$ 0 & \textbf{.99} $\pm$ 0 & \textbf{1.0} $\pm$ 0 & \textbf{.94} $\pm$ 0 && .999 $\pm$ 0 & .28 $\pm$ 0 & .99 $\pm$ 0 & \textbf{1.0} $\pm$ 0 & .82 $\pm$ 0 && \textbf{.87} $\pm$ 0 & \textbf{.83} $\pm$ 0\\ 
    \phantom{A} + LSTM  & .98 $\pm$ 0& 0 $\pm$ 0& 0 $\pm$ 0& 0 $\pm$ 0& .25 $\pm$ 0&& \textbf{1.0} $\pm$ 0& 0 $\pm$ 0& 0 $\pm$ 0 & 0 $\pm$ 0& .25 $\pm$ 0 && .25 $\pm$ 0 & 0 $\pm$ 0\\
    \phantom{A} + TCN  & .99 $\pm$ .002 & 0 $\pm$ 0& .28 $\pm$ .65 & .08 $\pm$ .03 & .34 $\pm$ .17  && .81 $\pm$ .55 & 0 $\pm$ 0 & .84 $\pm$ .32 & .51 $\pm$ .37 & .57 $\pm$ .11 && .35 $\pm$ .13 & .15 $\pm$ .17\\ 
    \phantom{A} + Transformer  & \textbf{.999} $\pm$ .001 & .033 $\pm$ .007 & .49 $\pm$ .22 & .11 $\pm$ .01 & .41 $\pm$ .05 && .80 $\pm$ .05 & \textbf{.89} $\pm$ .12 & .999 $\pm$ .003 & .96 $\pm$ .06 & \textbf{.91} $\pm$ .02 && .45 $\pm$ .06 & .30 $\pm$ .07\\
    \bottomrule
    \end{tabular}
    \label{tab:results}
    \end{center}
\end{table*}

\textbf{The neuro-symbolic model has the best overall performance.} The Neural \textit{AE} + FSM model achieves the best \textit{precision} and $F1$ scores, and reasonably good \textit{recall} scores. However, an exception happens for $e_1$. It is possible that a classification mistake made on an \textit{AE} such as ``\textit{wash hands}" and ``\textit{walk}" directly leads to an incorrect detection of $e_1$ compared to other \textit{CE}s. Hence, the accuracy of the \textit{AE} classifier plays a pivotal role in the performance of this model.

\begin{figure}
    \centering
    \includegraphics[width=1\columnwidth]{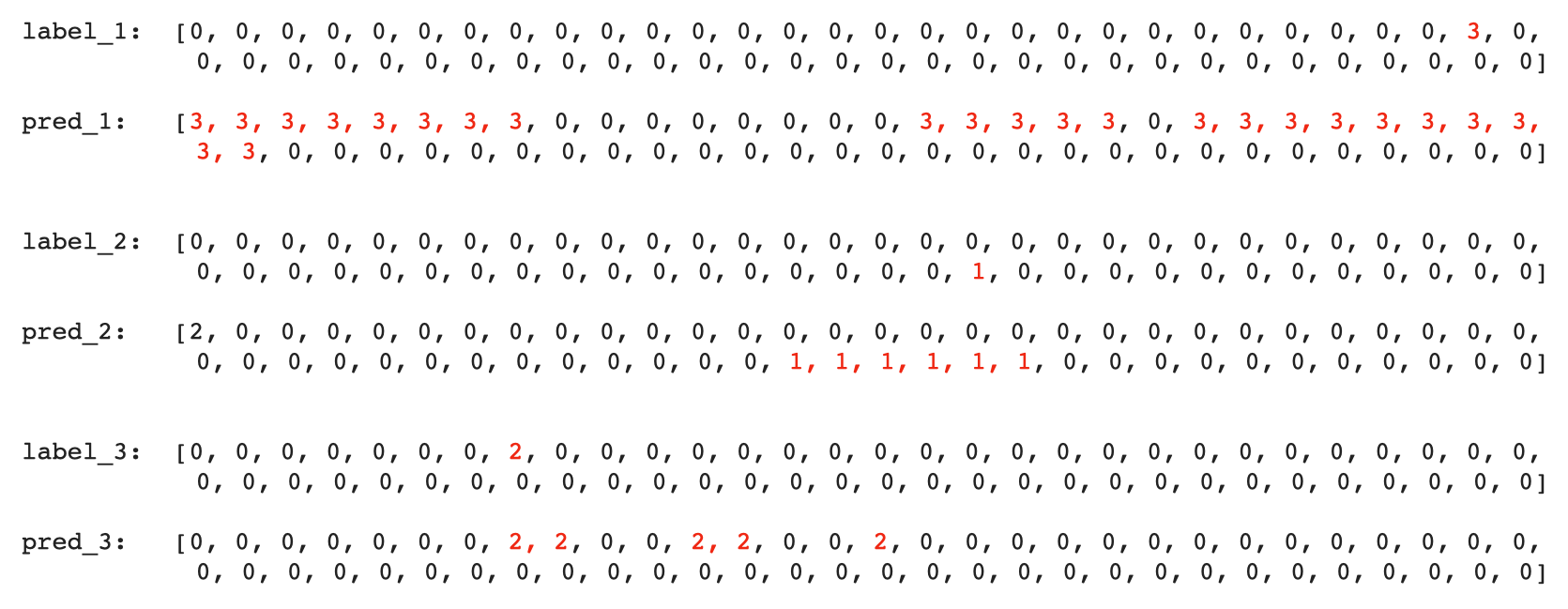}
    \caption{Example \textit{CE} label sequences predicted by NN models. Positive \textit{CE} labels are highlighted in red in both ground-truth  (\texttt{label\_i}) and corresponding prediction (\texttt{pred\_i}).}
    \label{fig:examples}
\end{figure}

\textbf{Tranformer and Neural \textit{AE} + Tranformer are best among NNs.} While LSTMs theoretically have infinite memory, they perform the worst. The TCN model that takes 128-dim embedding is better than the Neural \textit{AE} + TCN that uses 9-dim one-hot \textit{AE} vectors, but for Transformer and Neural  \textit{AE} + transformer models, things are the opposite. This may indicate TCNs are better at analyzing high-dim noisy data, while Transformers perform well on already extracted symbolic \textit{AE} concepts. \textit{AE} + Tranformer, Tranformer and TCN models achieve relatively high \textit{recall} scores for each class, indicating their ability to identify the occurrence of complex events. \textit{AE} + Tranformer and Tranformer have the highest $F1$ scores among NNs. However, NN models exhibit low \textit{precision} scores for all positive \textit{CE} classes, suggesting that they struggle to capture the precise timing of each complex event. This limitation becomes more evident when examining their predicted label sequences in Figure~\ref{fig:examples}. Rather than generating a single positive \textit{CE} label, the NNs tend to produce consecutive positive labels. 

\subsubsection{Evaluation of NNs with various training dataset sizes}
\begin{figure}
    \centering
    \includegraphics[width=0.75\columnwidth]{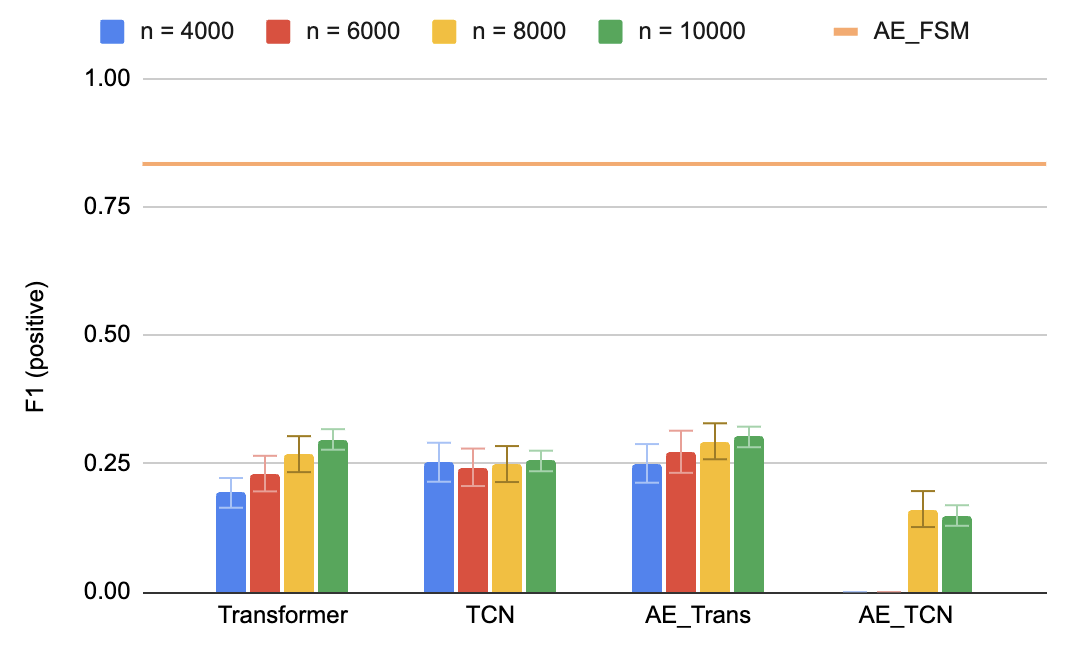}
    \caption{Evaluation of NN models with various \textit{CE} training data sizes.}
    \label{fig:train-size}
\end{figure}

To study the influence of \textit{CE} training dataset sizes on the performance of NNs, we create datasets with 8,000, 6,000, and 4,000 examples. The results of training the Transformer, TCN, Neural \textit{AE} + Transformer, and Neural \textit{AE} + TCN on various dataset sizes are shown in Fig.~\ref{fig:train-size}.

\textbf{Trasnformer models have the potential to perform better with more data.} Despite being trained on a larger dataset, the TCN model's performance plateaus without further improvement, while Transformer and Neural \textit{AE} + Transformer models show an upward trend in performance. However, they are still far worse than the overall performance of the neuro-symbolic model. This is because the FSMs in the Neural \textit{AE} + FSM model already contain human knowledge of \textit{CE} patterns, saving the efforts to learn from a great amount of data.

\section{Conclusion}
Our proposed methodology enables the synthesis of stochastic CE datasets, facilitating fair performance comparison between NNs and NS architectures. Our evaluation demonstrates that, in general, NS methods greatly outperform NNs due to the injection of human knowledge. Although some neural-only architectures can further improve as training data sizes increase, they have limited context sizes and are unsuitable for real-life complex events that span hours. We also observed that the same model can exhibit different performance levels in detecting various complex events. It would be advantageous for future works to quantify the complexity of complex events to benchmark models.

\addtolength{\textheight}{-12cm}   





\section*{Acknowledgement}
The research reported in this paper was sponsored in part by the DEVCOM Army Research Laboratory (cooperative agreement \# W911NF1720196 ), the Air Force Office of Scientific Research (award \# FA95502210193), the National Institutes of Health (award \# 1P41EB028242), and the National Science Foundation (NSF) under award \#1822935. The views and conclusions contained in this document are those of the authors and should not be interpreted as representing the official policies, either expressed or implied, of the funding agencies.


\bibliographystyle{IEEEtran}
\bibliography{IEEEabrv,mybibfile}

\end{document}